\documentclass[conference]{IEEEtran}
\usepackage{cite}
\usepackage{amsmath,amssymb,amsfonts}
\usepackage{algorithmic}
\usepackage{graphicx}
\usepackage{textcomp}
\usepackage{xcolor}
\def\BibTeX{{\rm B\kern-.05em{\sc i\kern-.025em b}\kern-.08em
    T\kern-.1667em\lower.7ex\hbox{E}\kern-.125emX}}

\usepackage{subcaption}
\usepackage{ctable}
\usepackage{bbold}

\graphicspath{{figures/}}

\newcommand*{\Eqref}[1]{Eq.~\eqref{#1}}
\def\*#1{\mathbf{#1}}

\begin{document}

\title{Recurrent Point Review Models}

\author{\IEEEauthorblockN{Kostadin Cvejoski\IEEEauthorrefmark{1}\IEEEauthorrefmark{3}, Rams\'es J. S\'anchez\IEEEauthorrefmark{1}\IEEEauthorrefmark{2}, Bogdan Georgiev\IEEEauthorrefmark{3}, Christian Bauckhage\IEEEauthorrefmark{3} and C\'esar Ojeda\IEEEauthorrefmark{4}}
\IEEEauthorblockA{\IEEEauthorrefmark{1}Competence Center Machine Learning Rhine-Ruhr}
\IEEEauthorblockA{\IEEEauthorrefmark{2}B-IT, University of Bonn, Bonn, Germany}
\IEEEauthorblockA{\IEEEauthorrefmark{3}Fraunhofer Center for Machine Learning and Fraunhofer IAIS, 53757 Sankt Augustin, Germany}
\IEEEauthorblockA{\IEEEauthorrefmark{4}Berlin Center for Machine Learning and TU Berlin, 10587 Berlin, Germany\\
\{kostadin.cvejoski, bogdan.georgiev, christian.bauckhage\}@iais.fraunhofer.de,\\ojeda.marin@tu-berlin.de, sanchez@bit.uni-bonn.de}}

\maketitle

\begin{abstract}
Deep neural network models represent the state-of-the-art methodologies for natural language processing. 
Here we build on top of these methodologies to incorporate temporal information and model how review data changes with time.
Specifically, we use the dynamic representations of recurrent point process models,
%
%
which encode the history of how business or service reviews are received in time, 
to generate instantaneous language models with improved prediction capabilities. 
Simultaneously, our methodologies enhance the predictive power of our point process models by incorporating summarized review content representations. 
%
%
%
We provide recurrent network and temporal convolution solutions for modeling the review content.
We deploy our methodologies in the context of recommender systems, 
%
effectively characterizing the change in preference and taste of users as time evolves. Source code is available at \cite{source_code}.
\end{abstract}


\section{Introduction}
Dynamic models of text aim at characterizing temporal changes in patterns of document generation. Most successful dynamic language models are Bayesian in nature, and lag behind state-of-the-art deep language models in terms of expressibility. A natural space to study some of the temporal aspects of language is that of the large review datasets found in e-commerce sites. 
The availability of millions of reviewed items, such as business or services, books or movies, whose reviews have been recorded in time scales of years, opens up the possibility to develop deep scalable models that can predict the change in taste and preference of users as time evolves. Originally, the interaction of users in these e-commerce sites were studied in the context of collaborative filtering, where the goal was to predict user ratings, based on user interaction metrics. Here we aim to look directly at the content of reviews as time evolves.

Costumer reviews provide a rich and natural source of unstructured data which can be leverage to improve recommender system performance \cite{liu2019daml}. Indeed, reviews are effectively a form of recommendation.
Recently, a variety of deep learning solutions for recommendation have profit from their ability to extract latent representations from review data, encoding rich information related to both users and items.
%
%
Time represents yet another dimension of context, as user preference and item availability change with time
-- and indeed,
causal and temporal relations have been known to improve the performance of recommender systems \cite{wu2017recurrent} \cite{sachdeva2019sequential}. 
Despite this fact,
recent natural language processing (NLP) methodologies for rating and reviews \cite{zheng2017joint} lag behind at incorporating temporal structure in their language representations. In the present work we exploit recurrent neural network (RNN) models for point processes, and feed them neural representations of text, to characterize costumer reviews. Our goal is to capture the changes in user taste and item importance during time, and to exploit those changes to better predict when are new reviews arriving, and what do they actually say.
We summarize our contributions as follows:
\begin{itemize}
    \item A text-augmented marked temporal point process model for review arrival times with better predictive performance,
    \item A hierarchical language model that leverages the global dynamic representations of the review arrival point process, effectively defining instantaneous language models with improved modelling performance.
\end{itemize}{}


\begin{figure}
 \includegraphics[width=.93\linewidth]{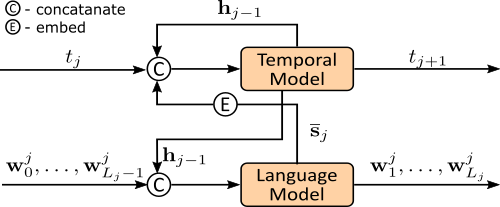}
 \caption{Recurrent Point Review (RPR) Model. We treat the pair $(t_j, (\mathbf{w}^j_0,\cdots\mathbf{w}^j_{L_j-1}))$ as input to our model. The hidden state of the temporal model $\mathbf{h}_j$ captures the non-linear dependency between timing and text content from past reviews. On the other hand, the mean hidden state of the language model $\bar{\mathbf{s}}_j$ holds information about the text content of the current review and is used for updating the global representation $\mathbf{h}_j$. See Table \ref{tab:notation_summary} for reference to our notation.}
\label{fig:rpr_language_model}
\end{figure}

We present the related work in Section \ref{sec:RW} and introduce our model in Section \ref{sec:Model}. The baseline models used for comparison in this paper are presented in Section \ref{ssec:BM}. The experimental setup and results are presented in Section \ref{sec:ER}. Finally, in Section \ref{sec:CF} we conclude and discuss future work.

\section{Related Work}
\label{sec:RW}

The dynamics of language is of fundamental importance in social sciences as a proxy for cultural evolution \cite{kirby2007innateness}. Complex system methods seek to understand the emergence of the encoding capabilities of language \cite{nowak1999evolution}, and evolutionary approaches
 -- following the Bayesian tradition from Phylogenetics, 
 study the competition between grammar and syntax in the context of historical linguistics \cite{greenhill2017evolutionary}. Closer to our line of work, research of online communities point to temporal linguistic changes as means to enforce community norms \cite{danescu2013no}. Our methodologies aim at studying similar systems in the e-commerce review context, wherein linguistic change is relevant in time scales of months and years.

The work on language dynamics from the machine learning community has mainly focused on the dynamics of word embeddings and topics. On the one hand, different embeddings, as e.g. word2vec \cite{mikolov2013distributed}, are trained in slices of temporal data and alignments methods are performed a posteriori \cite{hamilton2016diachronic, kulkarni2015statistically}. The probabilistic framing of word embeddings has, in contrast, allowed for a stochastic-process point of view of embedding evolution \cite{rudolph2018dynamic}. 
On the other hand, within the dynamic topic modelling approach,
%
%
the parameters of models like Latent Dirichlet Allocation are defined to follow known stochastic processes in time \cite{wang2012continuous,yogatama2014dynamic}. 
Likewise, self-exciting point processes in time have allowed for clustering of document streams \cite{du2015dirichlet,he2015hawkestopic}. Lastly, while writing this paper we found \cite{dynamic_language_model_}, in which a RNN language model is conditioned on a global dynamic latent variable.
%
In contrast to this work, our dynamic representations explicitly encode both timing and content of past reviews, and can capture non-Markovian dynamics.

%
Finally, within the recommender system realm deep neural networks models of review data for rating predictions use embedding representations, as well as convolutional neural networks \cite{catherine2017transnets}. They also provide characterization of review usefulness\cite{fan2019product}, use reviews for product description \cite{novgorodov2019generating}, and provide better representations for rating prediction \cite{esmaeili2019structured}. The need to interact with the costumer has also led to question answering solutions \cite{chen2019driven,yu2018aware}. Different from these works, we focus on the temporal aspects of review content.

\section{Recurrent Point Review Model (RPR)}
\label{sec:Model}

Rather intuitively, the generation process underlying the content of the reviews received by a business or service is intimately related to how fast and when these reviews are created (or received) in time. It is nor hard to imagine, for example, reviews with similar content clustering and forming different kind of patterns in time. Based on this intuition, we develop a model that explicitly uses the text content of reviews to better predict the arrival times of new ones. Simultaneously, the model leverages the history of how reviews have arrived in time to yield \textit{instantaneous} language models, i.e. language models which are informed by time. As a result we improve the text prediction capabilities of standard language models by learning representations which characterize the time evolution of the review content describing the business or service in question. 

Consider an item  $a$ (e.g. a business, service or movie) and assume that, since its opening to the public, it has received a collection of $N_a$ reviews $\*r_j^a = \{(\*x^a_{j},t^a_j)\}_{j=1}^{N_a}$, where $t^a_j$ labels the creation time of review $\*r^a_j$ and $\*x^a_j = (\*w^{a,j}_1,...,\*w^{a,j}_{L_{a,j}})$ corresponds to its text\footnote{In what follows we shall drop the index $a$ over items.}. Such a collection of reviews effectively defines a point process in time.
Our main idea is to model these point processes as RNNs \textit{in continuous time}, and feed them representations summarizing the content of past reviews. We then use the point process' hidden representations, which encode the nonlinear relations between text and timing of past reviews, to predict both (i) how the reviews' content of a given item changes with time, and (ii) when are new reviews going to arrive. The model thus consists of two interacting components: a neural point process model which leverages the information encoded in the review content, and a dynamic neural language model which uses the point process history. In what follows we dwell into the details of these two building blocks. Figure \ref{fig:rpr_language_model} summarizes the Recurrent Point Review model.

\begin{table}[t]
    \centering
    \begin{tabular}{ ll } 
        \textbf{Symbol}  & \textbf{Description}\\ 
        \specialrule{.1em}{.05em}{.05em} 
        $V$ & vocabulary size  \\
        $H$ & dimension of hidden state for temporal dynamics\\
        $S$ & dimension of hidden state for language model\\
        $W$ & word embedding dimension\\
        $a$ & item index (business)  \\
        $N_a$ & number of reviews for business $a$  \\
        $M$ & total number of businesses \\
        $\*r^a_j$ & review of business $a$ at time step $j$  \\
        $\mathbf{x}^a_j$ & text of review $\*r^a_j$ \\
        $L_{j}$ & number of words in the $j$th review \\
        $t_j^a$ & timestamp of review $\*r^a_j$ \\
        $\mathbf{w}_i^{j}$ & $i$th word embedding in the $j$th review \\
        $\mathbf{X}^a_j$ & bag of words embedding for review $\* r^a_j$ \\
        $\mathbf{h}^a_j$ & hidden state of the temporal model for review $\* r^a_j$ \\
        $\bar{\mathbf{s}}_j$ & summary content representation for the $j$th review \\
    \hline
    \end{tabular}
     \caption{Summary of notation used throughout the paper.}
    \label{tab:notation_summary}
\end{table}
%
%
%
%

\subsection{Dynamic language models for review text content} 
\label{ssec:LM}

Auto-regressive neural language models approximate the joint probability distribution over sequences of words with a product over conditional probabilities such that
\begin{equation}
p(\*x) = \prod^{L}_{i=0} \, p_{\theta}(\mathbf{w}_i|\mathbf{w}_{<i}),
\label{eq:LM_base}
\end{equation}
where $\*x =(\*w_1,...,\*w_{L})$ labels the sequence of words in question, 
 and $p_{\theta}$ is a discrete probability distribution, parametrized by a neural network with parameter set $\theta$, which yields the probability of observing $\*w_i$ given the previously generated words $\*w_{<i}$ \cite{Mikolov}. 
In order to capture how the review content of a given item changes with time, we assume the conditional probabilities in Eq. \eqref{eq:LM_base} additionally depend on a vector representation $\*h_j$, which encodes the nonlinear relations between creation times and content of past reviews. Let us assume for the moment that $\*h_{j-1}$ is given. We shall explain how to calculate it in Section \ref{ssec:RMPP} below. Given $\*h_{j-1}$, we can rewrite Eq. \eqref{eq:LM_base} for the $j$th review of an item thus
\begin{equation}
p(\*x_j|\*h_{j-1}) = \prod^{L_{j}}_{i=0} \, p_{\theta}(\mathbf{w}^{j}_i|\mathbf{w}^{j}_{<i}, \*h_{j-1}).
\label{eq:LM}
\end{equation}
In practice we define $p_{\theta}$ as a categorical distribution over a vocabulary of size $V$ whose class probabilities $\boldsymbol{\pi}^j_i \in \mathbb{R}^V$ are given by
\begin{equation}
    \boldsymbol{\pi}^j_i = \mbox{softmax}(\*W \, \*s^j_i), \quad \*s^j_i = g_{\theta}(\mathbf{w}^{j}_{<i}, \*h_{j-1}),
    \label{eq:hidden_state_lm}
\end{equation}
where the index ``$i$" runs from one to $L_j$, $\*W \in \mathbb{R}^{V \times S}$ is a learnable weight matrix, and $\*s_i^j \in \mathbb{R}^S$ is the hidden representation of the $i$th word in review $j$. We shall model the function $g_{\theta}$ with two standard neural network architectures, namely a Long short-term memory (LSTM) neural network and a Temporal Convolutional Network (TCN). We briefly revisit these two architectures below.
\subsubsection{Long Short Term Memory Network}
\label{ss:LSTM}
One of the most common variants of recurrent neural networks, aimed at solving the vanishing gradient problem, is the long short-term memory network \cite{LSTM}. Let us define $\hat{\*w}_i^j = \mbox{concat}(\*w_i^j, \*h_{j-1})$. The LSTM network recursively processes each element $\hat{\*w}_i^j$ in review $j$ while updating its hidden state $\*s^j$ as follows
\begin{figure}[t]
    \centering
    \begin{subfigure}[ht]{0.40\textwidth}
         \centering
         \includegraphics[width=\linewidth]{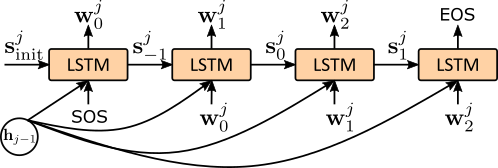}
         \label{fig:language_rnn_model}
     \end{subfigure}
     \par\bigskip
    \begin{subfigure}[ht]{0.3\textwidth}
         \centering
         \includegraphics[width=\linewidth]{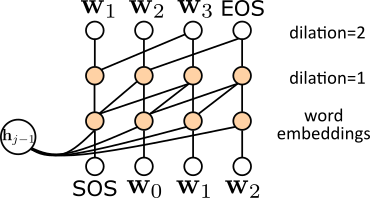}
         \label{fig:language_cnn_model}
     \end{subfigure}
    \caption{Neural networks parametrizing our dynamic language models. 
    \textit{Upper figure}: LSTM network for which the hidden state $\*s^j_i$ is updated recursively. \textit{Lower Figure}: Dilated TCN, whose effective receptive size grows exponentially with each layer. The temporal vector representation $\mathbf{h}_{j-1}$ is concatenated with each word embedding in both models. S0S and EOS label the star-of-sequence and end-of-sequence tokens. See Table \ref{tab:notation_summary} for reference to our notation.}
    \label{fig:lm_models}
\end{figure}
\begin{align}
\label{eq:lstm_cell}
\*i_t & =  \sigma(\*W^1_i \, \hat{\*w}_t + \*W^2_i \,  \*s_{t-1}+\* b_i), \nonumber \\
\*o_t & =  \sigma(\*W^1_o \, \hat{\*w}_t + \*W^2_o \,  \*s_{t-1}+\* b_o), \\
\*f_t & =  \sigma(\*W^1_f \, \hat{\*w}_t + \*W^2_f \,  \*s_{t-1}+\* b_f), \nonumber\\
\*c_t & =  \*f_t \odot \*c_{t-1} + i_{t} \odot \mbox{tanh}(\*W^1_c \, \hat{\*w}_t + \*W^2_c \,  \*s_{t-1}+\* b_c), \nonumber\\
\* s_t & =  \* o_t\odot \mbox{tanh}(\*c_t). \nonumber
\end{align}
Here $\*i_t$, $\*o_t$ $\*f_t$, $\*c_t$, and $\*s_t$ are the input, output, forget, memory and hidden states of the LSTM, respectively; ``$t$" runs from one to $L$, the number of words in the $j$th review (note we have omited $j$ above); $\sigma$ labels the ReLU nonlinearity and  $\odot$ labels element-wise multiplication. The upper sub-figure in Fig. \ref{fig:lm_models} depicts the LSTM dynamic language model.

\subsubsection{Temporal Convolutional Network}
\label{ss:TCN}
Convolutional neural networks have shown to be the workhorse for image processing tasks \cite{lecun1998gradient, lecun2015deep}, and NLP applications have profit from their rich representations of text \cite{denil2014modelling}. To extend these model architectures for sequential modelling tasks, one must enforce two conditions \cite{bai2018empirical}:
(i) no information “leakage” from the future to the past, and (ii) the architecture should take a sequence of any length and map it to an output sequence of the same length. Both these conditions can be achieved with the correct padding, and we refer to \cite{bai2018empirical} for reference.

\textit{Dilation}: One would like the network to look very far into the past. This characteristic is accomplished by what is known as diluted convolutions \cite{yu2015multi}. Essentially, dilation boils down to skipping $d-1$ inputs at each step. More formally, for a filter $f : \{0,...,k-1\}\rightarrow \mathbb{R}$, the dilated convolution operation $g$ on the $i$th word of review $j$ reads
\begin{equation}
g_{\theta}(\*w^j_i) = \sum^{k-1}_{l=0}f_{\theta}(l)\cdot \*x^j_{i-dl},
\label{eq:dilated}
\end{equation}
\begin{figure}[t]
     \centering
     \includegraphics[width=0.3\linewidth]{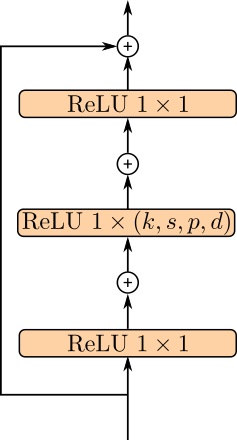}
     \caption{Temporal Convolution Network: Residual Block}
     \label{fig:residual block}
    \label{fig:models}
\end{figure}
where $d$ is the dilation factor and $k$ is the filter size. With dilation, the effective receptive size can grow exponentially\footnote{In practice, we double the dilation step at each layer, i.e. $d_{l+1} = 2 d_l$.}.

\textit{Residual Block}: Our TCN model also uses residual connections \cite{resnet} to speed up convergence.
Specifically, we use a residual block similar to that of \cite{bytenet, DilatedCNNVAE}. The block is sketched in Fig. \ref{fig:residual block}.

The sequence of word representation $\*s^j_i$ of Eq. \eqref{eq:hidden_state_lm} is thus given by
\begin{equation}
    \*s^j_i = g_{\theta_l}(\mbox{ReLU}(g_{\theta_{l-1}}( \, \dots \, \mbox{ReLU}(g_{\theta_{1}}(\hat{\*w}_i^j))\dots))),
\end{equation}
where $l$ is the total number of layers in the TCN, $\hat{\*w}_i^j = \mbox{concat}(\*w_i^j, \*h_{j-1})$ and $g_{\theta}$ is defined in Eq. \eqref{eq:dilated} above. The lower sub-figure in Fig. \ref{fig:lm_models} shows the TCN dynamic language model.

\subsection{Temporal model for review creation: Recurrent Point Process (RPP) model}
\label{ssec:RMPP}

The collection of reviews received by a business or service effectively defines a point process in time. Let us consider a point process with compact support $\mathcal{S} \subset \mathbb{R}$. Formally, we write the likelihood of a new arrival (i.e. a new review) $\*r_{j+1}$ as an inhomogeneous Poisson process between reviews, conditioned on the history $\mathcal{H}_j \equiv  \{\*r_1,...,\*r_j\}$\footnote{a.k.a. filtration.} \cite{VereJones}. 
For the one-dimensional processes we are concerned here, the conditional likelihood function reads
\begin{equation}
f^*(t) = \lambda^*(t) \exp\left\{ \int^{t}_{t_j}\lambda^*\left(t'\right)dt' \right\}\, ,
\label{eq:point_likelihood}
\end{equation}
where $\lambda^*$ is (locally) integrable and is known as the intensity function of the point process.
Following \cite{NeuralHawkes}, \cite{RecurrentTemporal}, we define the functional dependence of the intensity function to be given by a RNN with hidden state $\*h_j \in \mathbb{R}^H$, where an exponential function guarantees that the intensity is \textit{non-negative}
\begin{equation}
\lambda^*(t) = \exp{\left\{\*v^t\cdot\*h_j + w^t\left(t-t_j\right) + b^t\right\}} \label{eq:intensity}\, .
\end{equation}
Here the vector $\*v^t \in \mathbb{R}^d$ and the scalars $w^t$ and $b^t$ are trainable variables. 

\begin{figure}[t]
     \centering
    \includegraphics[width=0.75\linewidth]{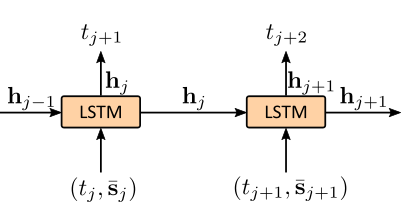}
    \caption{Text-augmented Recurrent Point Process (RPP) model. We treat the pair $(t_j, \bar{\mathbf{s}}_j)$ as input to the temporal model. The hidden state $\mathbf{h}_j$ captures the non-linear dependency between timing and content of the review arrival process. The summary representation of the language model $\bar{\mathbf{s}}_j$ holds information about the text content of the current review and is used for updating the global representation $\mathbf{h}_j$. See Table \ref{tab:notation_summary} for reference to our notation.}
    \label{fig:rpp_model}
\end{figure}

Now, we can modify the original RPP model of \cite{RecurrentTemporal} and augment $\*h_j$ to explicitly encode the nonlinear relations between content and timing of past reviews. Consider the summary representation $\bar{\*s}_j$ for review $\*r_j$ defined as the average 
\begin{equation}
    \bar{\*s}_j = \frac{1}{L_j}\sum_i^{L_j} \*s^j_i,
\end{equation}
where $\*s^j_i$ is given by Eq. \eqref{eq:hidden_state_lm}. The update equation for the hidden representation $\*h_j$ can then be written as 
\begin{equation}
\*h_j = g_{\theta}(t_{j}, \bar{\*s}_j, \*h_{j-1}),
\label{eq:transitionfunction}
\end{equation}
where $t_j$ labels the creation time of review $\*r_j$ and $\theta$ denotes the network's parameters. In practice, $g_\theta$ is implemented via an LSTM network with update equations similar to those in \Eqref{eq:lstm_cell}. To summarize, we use the summary representations $\bar{\*s}_j$ of the review content as marks in the recurrent marked temporal point process \cite{du2016recurrent}. Figure \ref{fig:rpp_model} sketches our text-augmented RPP model. 

Inserting \Eqref{eq:intensity} into \Eqref{eq:point_likelihood} and integrating over time immediately yields the likelihood $f^*$ as a function of $\*h_j$, that is 
\begin{multline}
f^*(t|\*h_j) =  \exp\left\{\*v^t \cdot \*h_j + w^t\left(t-t_j\right) \right.\\ \left.+ b^t + \frac{1}{w^t}\exp\left\{\*v^t\cdot \*h_j + b^t\right\} \right. \\ 
 \left. - \frac{1}{w^t}\exp\{\*v^t\cdot\*h_j + w^t\left(t-t_j\right) + b^t\} \right\}.
 \label{eq:likelihood_rpp}
\end{multline}


\textit{Prediction and sampling}. In order to use the RPP model for both prediction and sampling we require $p(T|H_j)$, the probability that the next review arrives at time $T$ given the previous history until the arrival of $\*r_j$. In the Appendix we calculate the latter and give expressions for both the average time of the next arrival and the inverse of the cumulative function of $p(T|H_j)$, which is required to sample the next review arrival time via inverse transform sampling.

\subsection{Recurrent point review model log-likelihood}
\label{ssec:likelihood_full_model}

We shall train our model using maximum likelihood. The complete log-likelihood of the RPR model can by written as
\begin{equation}
    \mathcal{L}=\sum_{a=1}^{M}\sum_{j=1}^{N_a}\Big\{\log f^*(\delta^a_{j+1}|\*h^a_j)+ \log p(\*x^a_j| \*h^a_{j-1})\Big\},
    \label{eq:loss}
\end{equation}
where $f^*$ is the conditional likelihood function of the RPP model defined in Eq. \eqref{eq:likelihood_rpp}$; \delta^a_{j+1} \equiv t^a_{j+1}-t^a_j$ denotes the inter-review time for item $a$; and $p(\*x^a_j| \*h_{j-1})$ labels the joint probability distribution over the sequence of words in $\*r_j^a$, as defined in Eq. \eqref{eq:LM}.

\section{Baseline Models}
\label{ssec:BM}

In this section we present a set of dynamic review models to be used as baselines\cite{cvejoski2019recurrent} in the experiment section below. These baselines will help us test the relevance of some of the properties of the RPR model, as well as of some of the hypothesis implicitly made while defining it. We shall replace the auto-regressive neural language models (LM) of Section \ref{ssec:LM} with a Bag of Words (BoW) Neural Text Model. This allows us to test, for example, whether the review word order (i.e. its grammar and distributed semantics) helps in modelling the arrival times of new reviews,
or if it is enough to simply know the presence of some key words to successfully perform this task. 
We also consider replacing the RPP model of Section \ref{ssec:RMPP} with a simpler LSTM model. Such a modification helps us test e.g.  whether the point-process nature of how reviews are received in time (e.g. its self-exciting properties) really plays a role when modelling the evolution of review content. 

In total we consider four combinations of dynamic review models: (i) RPP+(BoW) in which we replaced the autoregressive LM with a BoW model, (ii) t-LSTM+(RNN) and (iii) t-LSTM+(CNN), for which we replaced the RPP model with a simple LSTM, and (iv) t-LSTM+(BoW), for which both language and RPP models were replaced. Below we describe the BoW and t-LSTM models in detail.
\begin{figure*}[t]
     \centering
      \begin{subfigure}[ht]{0.245\textwidth}
         \centering
         \includegraphics[width=\linewidth]{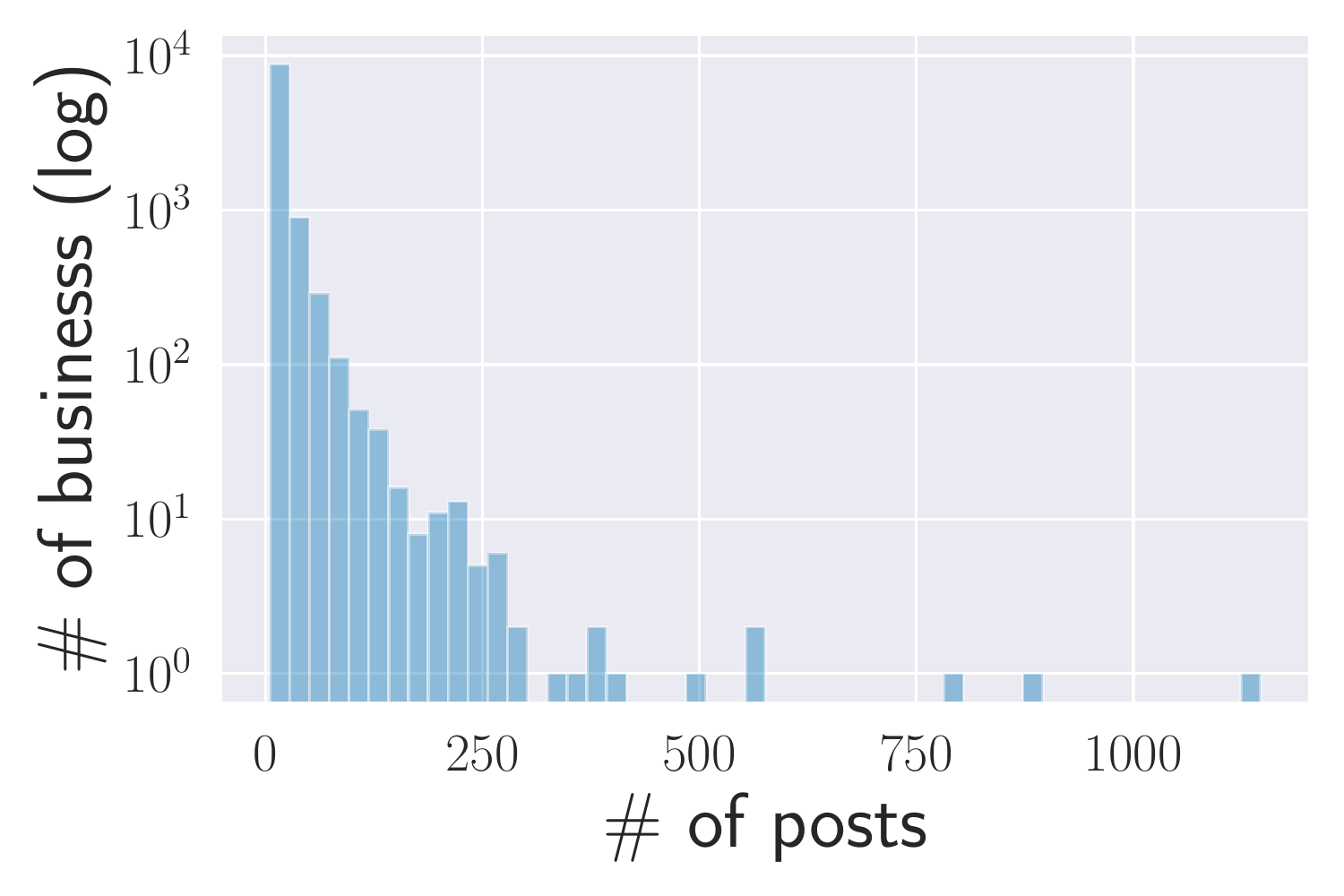}
         \caption{\# of arrivals per business}
         \label{fig:num_arrivals_per_business}
     \end{subfigure}
      \begin{subfigure}[ht]{0.245\textwidth}
         \centering
         \includegraphics[width=\linewidth]{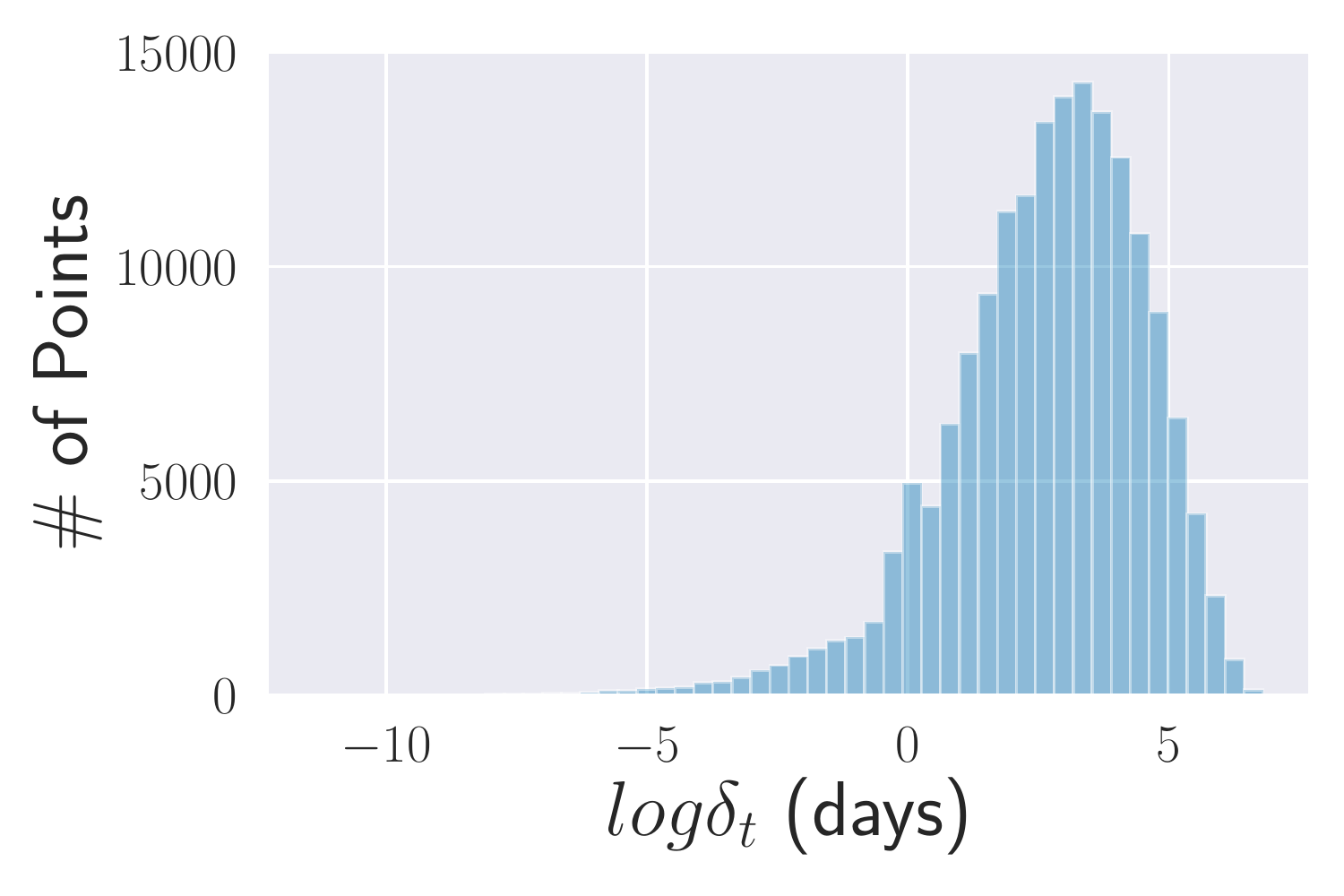}
         \caption{\textit{log} inter-arrival distribution}
         \label{fig:iterarrival_distribution}
     \end{subfigure}
          \centering
      \begin{subfigure}[ht]{0.245\textwidth}
         \centering
         \includegraphics[width=\linewidth]{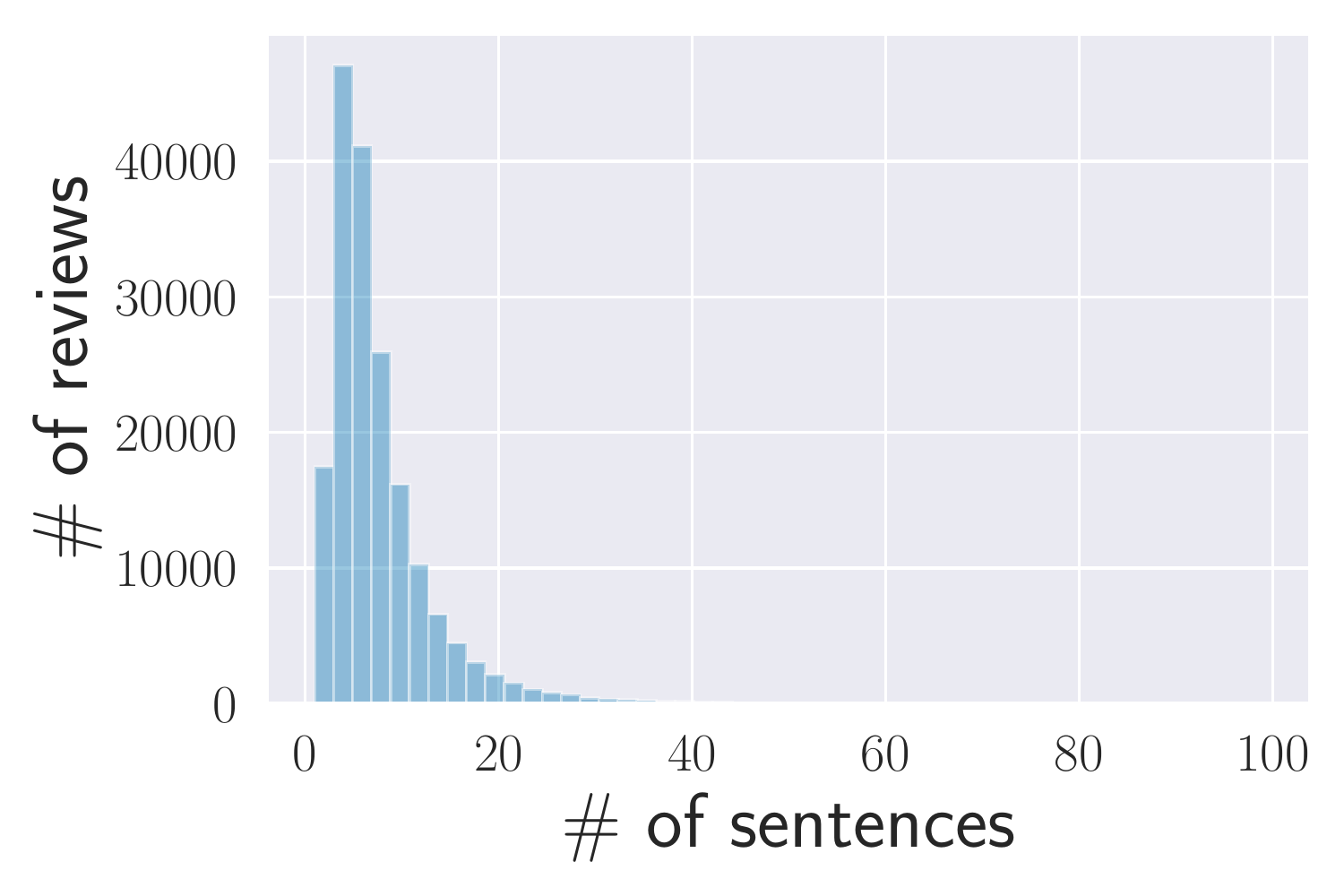}
         \caption{\# of sentences per review}
         \label{fig:num_sentences_per_review}
     \end{subfigure}
      \begin{subfigure}[ht]{0.245\textwidth}
         \centering
         \includegraphics[width=\linewidth]{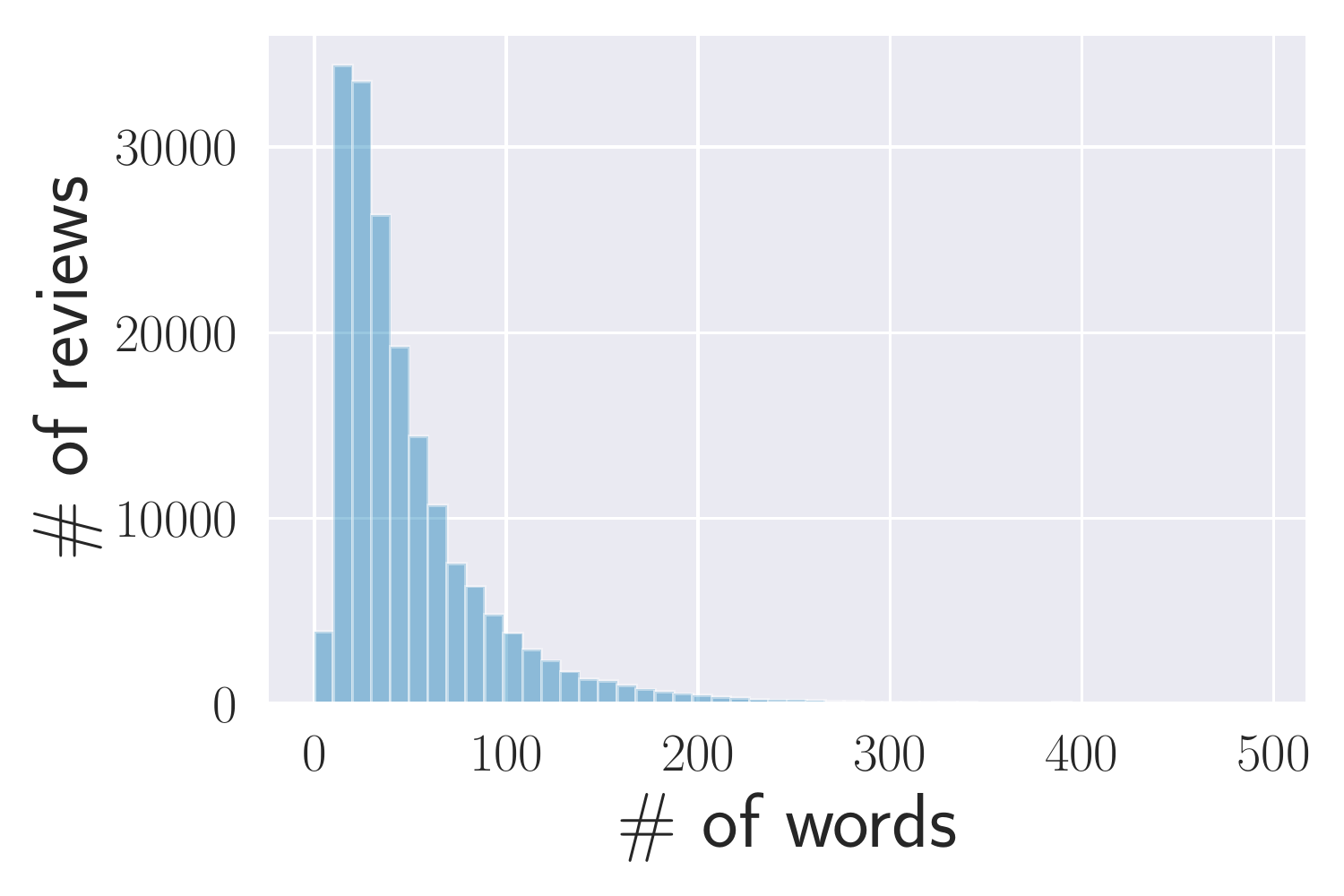}
         \caption{\# of words per review}
         \label{fig:num_words_per_review}
     \end{subfigure}
    \caption{Empirical distribution for different aspects of the data.}
    \label{fig:dataset_histograms}
\end{figure*}

\subsection{Bag of Words (BoW) Neural Text Model} 
\label{sssec:BoW}
We start by transforming the text of each review into a \textit{bag of words} (BoW) representation $\mathbf{X}_j\in \mathbb{R}^V$, where $V$ is the vocabulary size \cite{NIPS2009_3856}. To model the text component of reviews we assume the words in review $\*r_{j+1}$ are generated independently, conditioned on $\*h_j$, a vector representation describing the review creation dynamics. Note that $\*h_j$ can come from either the RPP model of Section \ref{ssec:RMPP} or the t-LSTM model of Section \ref{sssec:tLSTM}.
Following \cite{miao2016neural} we write the conditional probability of generating the $i$th word $\*w^{j+1}_i$ of the $(j+1)$th review as
\begin{align}
    p_\theta(\*w^{j+1}_i|\*h_j) =&  \frac{\exp{\{-z(\*w^j_i,\*h_j)\}}}{\sum_{k=1}^V \exp{\{-z(\*w^j_k,\*h_j)\}}},  \label{eq:word_prob} \\ 
    z(\*w^j_i,\*h_j) =& -\*h^T_j \*R \, \*w^j_i - \*b \, \*w^j_i, 
\end{align}
where $\*R \in \mathbb{R}^{H \times V}$ and $\*b \in \mathbb{R}^V$ are trainable parameters, and $\*w^{j}_i$ is the one-hot representation of the word at position $i$ of review $j$. 
We train this model via maximum likelihood. 

\subsection{t-LSTM model for review creation} 
\label{sssec:tLSTM}

The inter-review time $\delta_{j+1} \equiv t_{j+1}-t_j$ can be modelled as the mean of an exponential distribution with parameter $\lambda_{\theta}(\*h_j)$.  The function $\lambda_{\theta}$ is modelled by a multilayer perceptron and the hidden state encoding the history of review arrivals is given by
\begin{equation}
   \*h_j = f_{\theta}(t_{j}, \alpha_j, \*h_{j-1}),
\end{equation}
where $f_{\theta}$ is implemented by an LSTM network and $\alpha_j$ labels a text representation which can either be the summary representation $\bar{\*s}_j$ from the language models of Section \ref{ssec:LM} or the BoW representation $\*X_j$ of Section \ref{sssec:BoW}.

\begin{table}[t]
    \centering
    \begin{tabular}{ cccccc } 
        \textbf{Dataset} & \textbf{\#reviews} & \textbf{\#items} & \textbf{\#users} & \textbf{\#sentences} & \textbf{\#words} \\ 
        \specialrule{.1em}{.05em}{.05em} 
        Yelp-Shopping & 227K & 12K & 174K & 1,6M & 11,5M \\ 
    \hline
    \end{tabular}
     \caption{Yelp shopping category dataset statistics.}
    \label{tab:dataset_stats}
\end{table}

\begin{table}[ht!]
    \centering
    \begin{tabular}{ lrrrrrrrr } 
        &  \textbf{mean} & \textbf{std} &  \textbf{max} & \textbf{50\%}  & \textbf{95\%} & \textbf{99\%} \\ 
      \specialrule{.1em}{.05em}{.05em} 
      reviews & 17.57 & 29.99 & 1 147 & 10 & 53 & 123 \\ 
      sentences & 7.32 & 5.93 & 99  & 6 & 18 & 31 \\ 
      words & 50.41 & 46.40 & 493  & 36 & 136 & 238 \\ 
    \hline
    \end{tabular}
     \caption{Review statistics.}
    \label{tab:arrivals_stats}
\end{table}

\section{Experiments and Results}
\label{sec:ER}
\subsection{Data set}
For performance validation of the models we choose the Yelp19 dataset.\footnote{https://www.yelp.com/dataset}
We take all reviews for businesses that are labeled with the \textit{shopping} parent category from \textit{01 Jan 2016} to \textit{30 Nov 2018}. The creation time of a review is defined as the difference in days between the original timestamp and \textit{01 Jan 2016}. Next, we group reviews by business. All businesses with less than 5 reviews are removed. The sequential language models use the raw text from the reviews changed into lower case. In contrast, we convert the text from each review into a BoW vector of size 2000 \cite{NIPS2009_3856} for the BoW models. Preprocessing scripts can be found at \cite{source_code}.

The result from the preprocessing is summarized in Tables \ref{tab:dataset_stats} and \ref{tab:arrivals_stats}.  Note that the average number of reviews per business is 17.57, and 50\% of the businesses have less or equal to 10 reviews. The average length of a review is 50.41 sentences. This information gives insights helpful in the hyperparameter search during training. Finally, Figure \ref{fig:dataset_histograms} shows the number of posts per business distribution (Fig.  \ref{fig:num_arrivals_per_business}) and the \textit{log} inter-arrival distribution (Fig. \ref{fig:iterarrival_distribution}), as well as the number of sentences (Fig. \ref{fig:num_sentences_per_review}) and number of words (Fig. \ref{fig:num_words_per_review}) per review distributions.

\subsection{Training and Model Configuration}

\textit{Training.} The RPR models training on the objective \Eqref{eq:loss} has shown to be challenging. This is due to the fact that each model component (RPP and dynamic LM) use the latent representation from the other ($\bar{\mathbf{s}}_j$ and $\mathbf{h}_j$, respectively). Since we are using very flexible models, we have to be careful not to end up in a situation wherein one of the models is significantly much better than the other. Such a situation would cause one, or maybe both, of the models to ignore the latent embedding from the other one. To prevent such a situation, the frequency of gradient updates for each model should be tuned. Bear in mind that the RPP model is trained with \textit{backpropagation trough time} (BPTT) \cite{rumelhart1985learning,werbos1990backpropagation}, as to avoid the inherent RNN problems -- like vanishing or exploding gradient. Therefore, our propose solution is to run two forward steps for each BPTT window. In the first step, we freeze the parameters of the RPP model, and only update the parameters of the dynamic LM at each time step in the window, while storing the summary content representations. For the second forward run we freeze the parameters of the LM, and use the stored $\bar{\*s}_j$ as input to compute the gradients of the RPP model. This time we can perform gradient updates at the end of the BPTT window. Using one optimizer to train both models showed to be unstable.\footnote{This is due to the fact that optimizers like Adam \cite{ADAM} keep information about the dynamics of training. Using the same optimizer for the two models will lead to instabilities since the models have different learning dynamics.}

\textit{Model Configuration.} In our experiments, we randomly split each dataset into two parts: training set (80\%) and test set (20\%). We use grid search for hyper-parameters tuning. The temporal component of all dynamic review models have hidden state size of $H=128$, and we project $\bar{\mathbf{s}}_j$ into a 16-dimensional space before concatenation. The LM-RNN model has a hidden dimension $S$ of $1024$. We set the convolution filter to size $k=3$ for the LM-CNN model, and choose to use  6 residual blocks with dilation [1, 2, 4, 8, 16, 32], respectively. Weight normalization \cite{salimans2016weight} is used between the convolution layers. The vocabulary size is $V=3000$ for all sequential language models and we use the GloVe word embeddings \cite{pennington2014glove}, with dimension $W=300$.

We use Adam \cite{ADAM} to optimize the language model components of our models, with learning rate 0.0002 and $\beta_1=0.9$. The optimization of the temporal component was done with Adadelta \cite{zeiler2012adadelta}, initialized with a learning rate of 1. The size of the BPTT window for the RPP model is 20, since the average number of arrivals per business is 17.57. We limit the length of each review to be of 80 tokens. All methods are implemented using PyTorch v1.3\footnote{https://pytorch.org/}. Source code for all models can be found at \cite{source_code}.

\subsection{Results}

Given an item (business) of interest, the RPR model predicts both the arrival time of its next review and the probability of the word sequences within that review. To quantitatively evaluate the performance of the model on these tasks, we compute the root-mean-squared error (RMSE) $\varepsilon$ on the inter-review times, and the perplexity per word $\mathcal{P}$ with respect to the review content. Specifically, we define the RMSE as 
\begin{equation}
    \varepsilon = \sqrt{\sum_a^M \sum_j^{N_a} \frac{|\delta^a_j - \langle \tilde{\delta}^a_j \rangle |}{M N_a}},
\end{equation}{}
where $\langle \tilde{\delta}^a_j \rangle$ is the mean predicted inter-review time\footnote{To compute $\langle \tilde{\delta}^a_j \rangle$ via the RPP model we sample $t_j$ 1000 times using Eq. \eqref{eq:inverse_cum_rpp}. In the t-LSTM model $\langle \tilde{\delta}^a_j \rangle$ is obtained directly from $\lambda_{\theta}$.} and $\delta^a_j$ is its empirical value, $N_a$ is the number of reviews for business $a$ and $M$ the total number of businesses. Likewise we define the perplexity per word as
\begin{equation}
    \mathcal{P} = \exp\left\{-\sum_a^M \sum^{N_a}_{j}\sum_{i}^{L_{a, j}}\frac{\log p(\mathbf{w}^{a, j}_i| \mathbf{w}^{a, j}_{<i}, \*h^a_{j-1})}{M N_a L_{a, j}}\right\},
\end{equation}{}
where $\mathbf{w}^{a, j}_i$ and $L_{a, j}$ label the $i$th word and the number of words, respectively,  in the $j$th review of business $a$. 

To set yardsticks by which to judge the performance of our models, let us first analyse the review arrival process and the word distribution of the review content separately, that is, as though they were independent. Consider the time series defined by the review creation times $\{t_j^a\}_{j=1}^{N_a}$. We model the inter-review time of these series with (i) the RPP model of section \ref{ssec:RMPP} and (ii) the t-LSTM model of section \ref{sssec:tLSTM}, excluding in both cases any representation ($\bar{\*s}_j$ or $\*X_j$) encoding information from the reviews' content. The RMSE obtained via these models is shown in the first two rows of Table \ref{tab:results}. Note how the RPP model slightly outperforms the LSTM model. Likewise, consider the aggregated set of all review content received by all the businesses we study. We compute the perplexity per word of all reviews with respect to two language model, one parametrized by a LSTM and the other by a TCN. Let us refer to these models as LM-RNN and LM-TCN. Their perplexity is shown in the third and fourth rows of Table \ref{tab:results}. Note that for this data set the RNN performs better than the TCN. 

With these values as yardsticks we are now in position to discuss our results. Let us start by evaluating how well our dynamic review models predict the arrival time of new reviews. The results are shown in the second column of Table \ref{tab:results}. The first thing one can conclude from these results is that, as expected, review content does provide useful information for predicting when are the next reviews going to arrive. Even using a simple text representation as that of a BoW improves the RMSE values, as compared to the yardsticks. This makes sense intuitively, for one can imagine that reviews with similar content clusters in time. The last four rows of column 2 in Table \ref{tab:results} show our results for models whose content representation comes from auto-regressive LMs. All these instances beat the BoW models, which hints at the importance of accounting for word order. Finally our RPR models (with either RNN or TCN) improve by about 25$\%$ the RMSE values as compared to the yardsticks, and about 19$\%$ as compared to the baselines. This last comparison is surprising, given that RPP and t-LSTM models with no review content information yield similar RMSE values. Thus a marked point process model which uses the content summary representations from auto-regressive LMs as marks describes significantly better the empirical review arrival process, as compared to e.g. a stand alone RPP. 
\begin{table}[t]
    \centering
    \begin{tabular}{ lrrr } 
        \textbf{Model}  & \textbf{RMSE} $\downarrow$ & $\mathbf{R}^2$ $\uparrow$ & \textbf{Perplexity} $\downarrow$\\ 
        \specialrule{.1em}{.05em}{.05em} 
        t-LSTM & 96.8813  & 0.1788 & - \\
        RPP & 96.3794 & 0.1873 & - \\
        LM-RNN & - & - & 32.09 \\
        LM-TCN & - & - & 32.81 \\
        \hline
        t-LSTM+(BoW) & 95.3414 & 0.2046 & 519.90*\\
        RPP+(BoW)  & 92.3850 & 0.2533 & 511.32*  \\ 
        \hline
        t-LSTM+(RNN) & 89.9555 & 0.2920 & 29.86\\ 
        t-LSTM+(TCN) & 89.8053 & 0.2944 & 31.74\\ 
        RPR+(RNN)  & \textbf{71.8748} & \textbf{0.5480} & \textbf{29.61}  \\ 
        RPR+(TCN)  & 74.0945 & 0.5197 & 31.48  \\ 
        \hline
    \end{tabular}
     \caption{Model performance on RMSE, $\*R^2$ and perplexity per word. The RPR models significantly outperforms all baselines in all metrics. The perplexity for the BoW models corresponds to the predictive perplexity as defined in \cite{wang2012continuous}.
     }
    \label{tab:results}
\end{table}

Having seen that the review content information helps when modeling the review arrival process for the businesses we study, we can now ask whether the opposite flow of information is also useful, i.e.  does the dynamics of review arrival help in modeling the content of reviews? The last column in Table \ref{tab:results} shows the perplexity per word computed over all reviews received by a collection of businesses (i.e. aggregated over time). Low perplexity values indicate the model is able to predict well the sentences in a given review. Let us start with the BoW models. By definition, these models do not take into account the word sequence in a review $\*r_{j}$, and can only predict the distribution of words in the next review $\*r_{j+1}$. Accordingly, their perplexity values corresponds to the predicted perplexity, as defined in \cite{wang2012continuous}. Despite not being able to directly compare the BoW models with the LMs, we can see that RPP+(BoW) performs better than t-LSTM+(BoW). Let us move on now to our dynamic review models with summary content representation $\bar{\*s}_i$. All models yield lower perplexity values than the yardsticks. Therefore all models are able to successfully use the vector representation $\*h_j$ encoding the review arrival dynamics. We can conclude that its inclusion does make sequence words prediction easier. Note, in particular, that the RPR+(RNN) model outperforms all baselines. 
Finally, we have also computed the instantaneous perplexity, or perplexity per review, and noticed that the RPR models consistently yield better values than all baselines.


\section{Conclusion and Future Work}
\label{sec:CF}
In this work we introduced neural dynamic language models of text for review data. We are able to leverage dynamic representations of point process models in language modelling tasks, and augment the point processes with text representations.
%
We showed that our approach improves performance on both content and arrival times prediction, as well as opens the door for \textit{dynamic generative language models}. Future work includes the implementation of attention mechanisms, as well as the inclusion of neural factorization machines aimed at predicting ratings values.

 \section{Appendix} 
We start by denoting with $p(T|\mathcal{H}_j)$ the probability that the next point arrives at $T$ given the previous history until the arrival of $\*r_j$ --- we require $p(T|\mathcal{H}_j)$ for both prediction and sampling. First, notice that the probability of no review arriving between $t_j$ and $t_j + \tau$ can be obtained as an integral over $p(T|\mathcal{H}_j)$, say
\begin{equation*}
\exp\left\{ - \int^{t_j+\tau}_{t_j}\lambda(t)dt\right\} = \int^{\infty}_\tau p(T|\mathcal{H}_j)dT \equiv G(\tau),
\end{equation*}
with $p(T|\mathcal{H}_j) = -\frac{dG(T)}{dT}$, where we used the Poisson distribution for zero arrivals in the first expression.  Solving for $G(\tau)$ we find
\begin{equation}
G(\tau) = \exp\left\{-e^{\alpha_j}\frac{1}{w^t}\left(e^{w^t\tau}-1 \right) \right\},
\label{eq:g-function}
\end{equation}
with $\alpha_j = \*v^t\*h_j + b^t$. The average time of the next arrival is then given by
\begin{equation}
\mathbb{E}[T] = \int^\infty_0p(T|\mathcal{H}_j)T \, dT = \int^\infty_0 G(T) \, dT.
\end{equation}
Finally, in order to sample the next arrival time one can use inverse transform sampling on $P(T)$. To this end one requires the inverse of the cumulative function of $P(T)$. We calculate the cumulative function thus
\begin{align}
F[p(T|\mathcal{H}_j)] &= \int^\tau_0p(T|\mathcal{H}_j)dT \nonumber \\
&= -\int^\tau_0\frac{dG(T)}{dT}dT = G(0) - G(\tau) 
\end{align}
whose inverse function then follows

\begin{align}
F^{-1}[p(T|\mathcal{H}_j)](y) = & \frac{1}{w^t} \Big(- \alpha_j + \log \left\{ w^t \left(\log \left\{\frac{-1}{y-1}\right\} \right.  \right. \nonumber \\
& + \left. \left. \left. \frac{e^{\alpha_j}}{w^t} \right) \right\} \right).
\label{eq:inverse_cum_rpp}
\end{align}

\section*{Acknowledgment}
The authors of this work were supported by the Fraunhofer Research Center for Machine Learning (RCML), within the Fraunhofer Cluster of Excellence Cognitive Internet Technologies (CCIT), and by the Competence Center for Machine Learning Rhine Ruhr (ML2R) which is funded by the Federal Ministry of Education and Research of Germany (grant no. 01|S18038A). Part of the work was also funded by the German Ministry for Education and Research as BIFOLD-Berlin Institute for the Foundations of Learning and Data (ref. 01IS18025A and ref 01IS18037A). We gratefully acknowledge this support.

\bibliographystyle{IEEEtran}
\bibliography{IEEEabrv,main}

\end{document}